# Unfinished Architectures: A Perspective from Artificial Intelligence

Elena Merino Gómez, Universidad de Valladolid, elena.merino.gomez@uva.es

Fernando Moral Andrés, Universidad Nebrija, fmoral@nebrija.es

Pedro Reviriego Vasallo, Universidad Politécnica de Madrid, pedro.reviriego@upm.es

***ABSTRACT***

Unfinished buildings are a constant throughout the history of architecture and have given rise to intense debates on the opportuneness of their completion, in addition to offering alibis for theorizing about the compositional possibilities in coherence with the finished parts.

The development of Artificial Intelligence (AI) opens new avenues for the proposal of possibilities for the completion of unfinished architectures. Specifically, with the recent appearance of tools such as DALL-E, capable of completing images guided by a textual description, it is possible to count on the help of AI for architectural design tasks. In this article we explore the use of these new AI tools for the completion of unfinished facades of historical temples and analyse the still germinal stadium in the field of architectural graphic composition.

## INTRODUCTION

The irruption of new applications of Artificial Intelligence seems to be covering all areas of knowledge with different possibilities of success. During the last decade it has experimented an exponential development, rapidly increasing the list of tasks in which it improves human performance. For example, the introduction of deep learning combined with the availability of large data sets for training has led to a remarkable improvement in the performance of artificial intelligence systems for vision and image processing applications (Elgendy, 2020). These applications include diagnosis from medical images (Barragán-Montero, 2021), detection of pedestrians by vehicles (Boukerche, 2021) or recognition of handwritten signatures, among many others. More recently, artificial intelligence tools are capable of generating images from text descriptions have emerged (Ramesh, 2021). These tools like DALL-E, MidJourney, Text2Art or Stable Diffusion are trained with hundreds of millions or billions of text/image pairs and use models with tens or hundreds of millions of parameters. This allows them to achieve high-quality results in many cases, although they still have limitations, for example, when the input text is in a language other than English or unusual words are used. Be that as it may, these new tools are opening up unprecedented possibilities in the world of visual arts.

However, as was the case in the 19th century, when architects believed they were capable of deducing the right solution for unfinished projects, today, the expectations created by image completion tools seem to be leading us down paths of similar enthusiasm. The nineteenth-century conviction that it was possible to propose the correct solution to the enigma of an unfinished project (Mangone, 2018, p. 10) seems to have a second chance with Artificial Intelligence. However, the very approach of "the correct solution" presupposes an unequivocal result, alien to the natural processes of image creativity, in which corrections and *pentimenti* have always been a common trend.

## *THE NEVER-ENDING STORY*

The fascination exerted by unfinished architecture is as old as the history of the world. The ὕβρις (hýbris) against the gods was sometimes resolved by the punishment consisting of the interruption of buildings. Thus, the Babylonians were scattered, leaving their tower (fig. 1) and their city uncompleted. The same misfortune befell the Olympion of Athens [1], whose incompleteness is attributed to the deific corrective of the tyrannical arrogance of the Pisistratids [2]. Despite the legendary and divine shadows that try to justify some inconclusions, economic, political, cultural or religious aspects are the ones that most frequently explain the huge amount of unfinished historical architecture that abounds throughout the entire European geography.

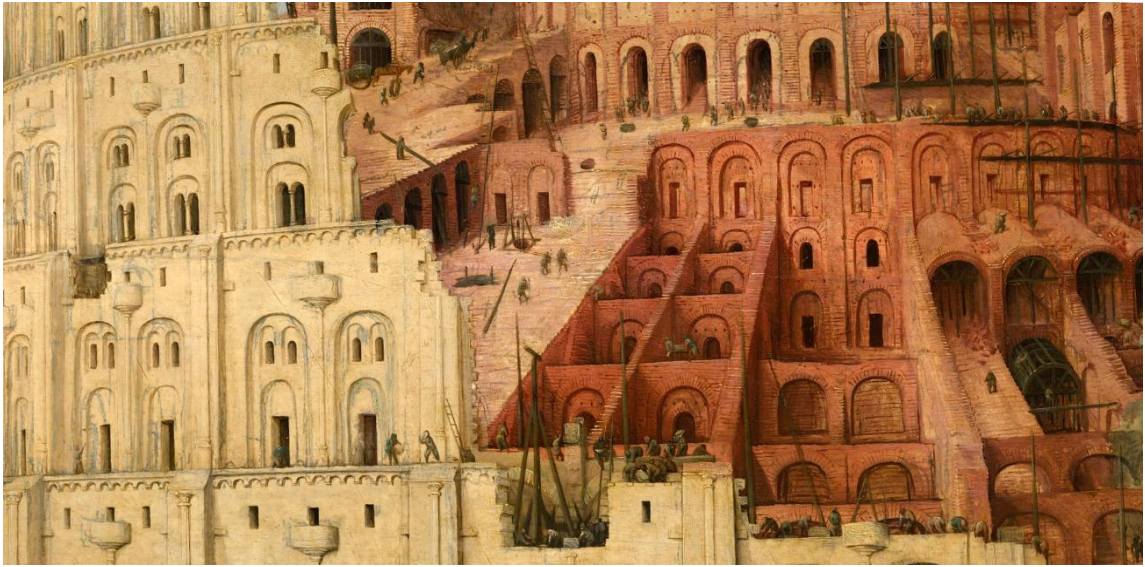

Fig. 1: *Tower of Babel.* (Detail). Pieter Brueghel, the Elder. 1563. Kunsthistorisches Museum, Vienna.

Sebastiano Serlio regretted in 1547, the year of the *editio princeps* of his Book V on the temples, the unfinished state of many of the great churches begun in Italy in his century and in previous ones. Huge buildings such as the Basilica of San Petronio, in his native Bologna, the Duomo in Milan, Santa Maria del Fiore in Florence or the ambitious work of San Pedro in Rome were being executed at that time. The constant threat of never reaching its completion loomed over them, either due to the customary mutability of economic resources or the fragile political balances:

> "[…] a´nostri tempi, o per la poca divotione, o per l´avaritia de li huomini, non si comincia più Chiesa che habbia del grande, né anche si finiscono le già cominciate." [3]

> (Serlio and Martin, 1547, p. 202)

The discussion about the advisability of resuming unfinished works of architecture shows as many edges as the etiology of its interruption. The economic reasons are the ones that are most frequently conjured up for the abandonment of the built work in general and of the façades, since generally superfluous, in particular.

Until the end of the 19th century and the beginning of the 20th, unfinished architectures were doomed to incompleteness. However, both public and specialized opinion were sensitive to the concept of glaring imperfection that an incomplete building conveyed. Of all the parts of a building, the façade is presented as a kind of maximum flagship of inconclusiveness, directly impacting the urban landscape, recalling old failed endeavours. Some projects for the completion of façades were the subject of frenzied controversies that transcended the professional field of architecture, involving large segments of the population to determine how they should be finished.

### *CORAM POPULO*

There is no architectural element more rigorously exposed to public opinion than the façade of a building. The perception of collective belonging makes façades a usual object of criticism and, among the non-specialized public, there is an unconscious identification between façade and architecture. Plebiscite events that offer citizens the opportunity to define their architectural or urban preferences rarely focus on the interior of buildings, but are relegated to the exterior.

Throughout the centuries, participatory processes of various kinds have marked the history, so often frustrated, of the completion of façades. The traditional competitions (Savorra, 2012), open or restricted, have been judged by the most diverse actors [4]: from the noble families that in thirteenth-century Florence voted for different solutions for Santa Maria del Fiore (Saalman, 1964, p. 490) to popular referendums, such as the one organized at the end of the 19th century for its definitive coronation (Franceschini, 1883), (fig. 2) or the one held in the 1930s for the conclusion of the façade of San Petronio in Bologna (Piacentini, 1935).

Once the historicist currents in the field of restoration were overcome in the first decades of the 20th century, the exercises for completing unfinished façades seem practically consumed. The first third of the 20th century witnessed the last competitions for the completion of historic architecture, strongly impregnated by ideological surroundings (Savorra, 2018), often unrelated to rigorous theoretical reasoning. Despite the definitive resignation of contemporary restoration trends for the completion of architectures in a historicist key, Matteo Renzi, mayor of Florence between 2009 and 2014, proposed in 2011 to complete the façade of San Lorenzo in Florence according to Michelangelo's project [ 5].

The positivism of the 19th century gives rise to the perception that, with the new instruments of analysis, architects would be able to complete buildings just as their authors would have done in their time if the work plans had not been frustrated (Mangone, 2018, p.10). The famous Viollet's maxim, states that "restoring a building is not maintaining, repairing or redoing it, it is restoring it to a complete state that may never have existed at a given time" [6] and is unequivocal in his conviction that no civilization, in times prior to his, would have intended to make restorations as they were understood in the second half of the 19th century [7]. The impossibility of "raising the dead" (Ruskin, 1889, p. 194) will support the scathing criticisms [8] with which his own contemporaries and his successors will attack him to the point of eclipsing his solid theoretical foundations.

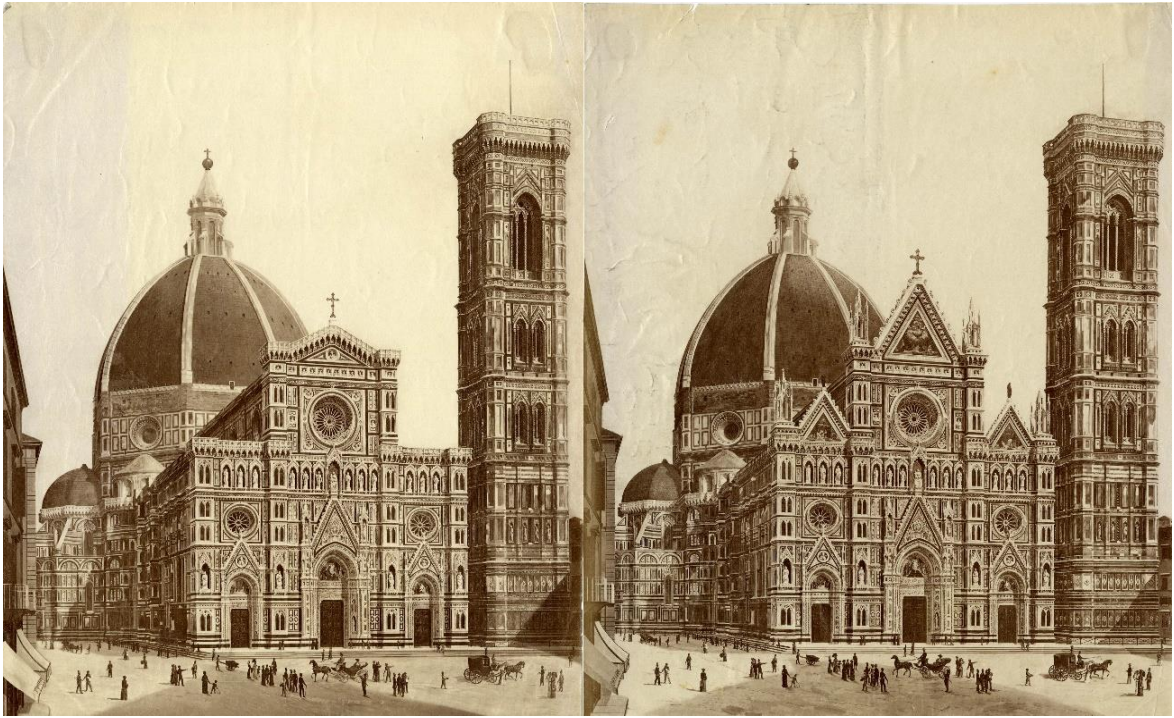

Fig. 2. Proposals for crowning the façade of the aisles submitted for consultation on January 3, 1884 (Zuffanelli and Faggia, 1887, p.16). On the left, the typical basilica crown model, on the right the tricuspid model. Niccolò Barducci, *Vedute di Santa Maria del Fiore with the facciata of the De Fabris*, 1876-83, Museo dell'Opera del Duomo di Firenze

### *IMAGINATION VS SCIENCE*

Despite the scientism ideas that sought a "correct" and therefore single solution, the disparate and sometimes crazy projects that were proposed for unresolved façades of some of the most important Italian churches confirm the infinity of solutions that can be proposed starting from a same precedent. In many cases, these are exercises of creative freedom, more or less based on the current trends in restoration in vogue at all times and on the rules of architectural composition. Even today, outside the scope of official competitions, solutions continue to be proposed, sometimes bordering on provocation. The above-mentioned example of the Basilica of San Petronio has been the subject of a termination project called *Bosco San Petronio* (figs. 3 y 4) brilliantly proposed by Mario Cucinella Architects (Totaro, 2020). In an even more conceptual and fanciful path are framed the Mario Mariotti's projections (1980) on the surface of the unadorned façade of the Santo Spirito in Florence (fig. 5).

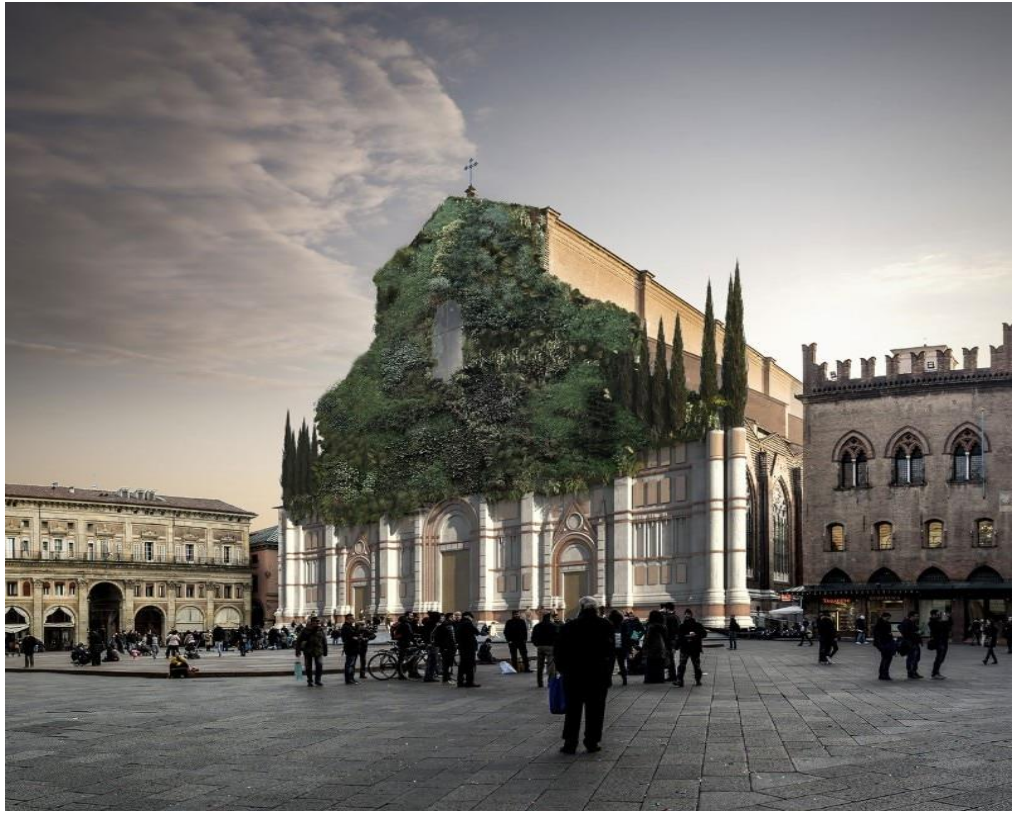

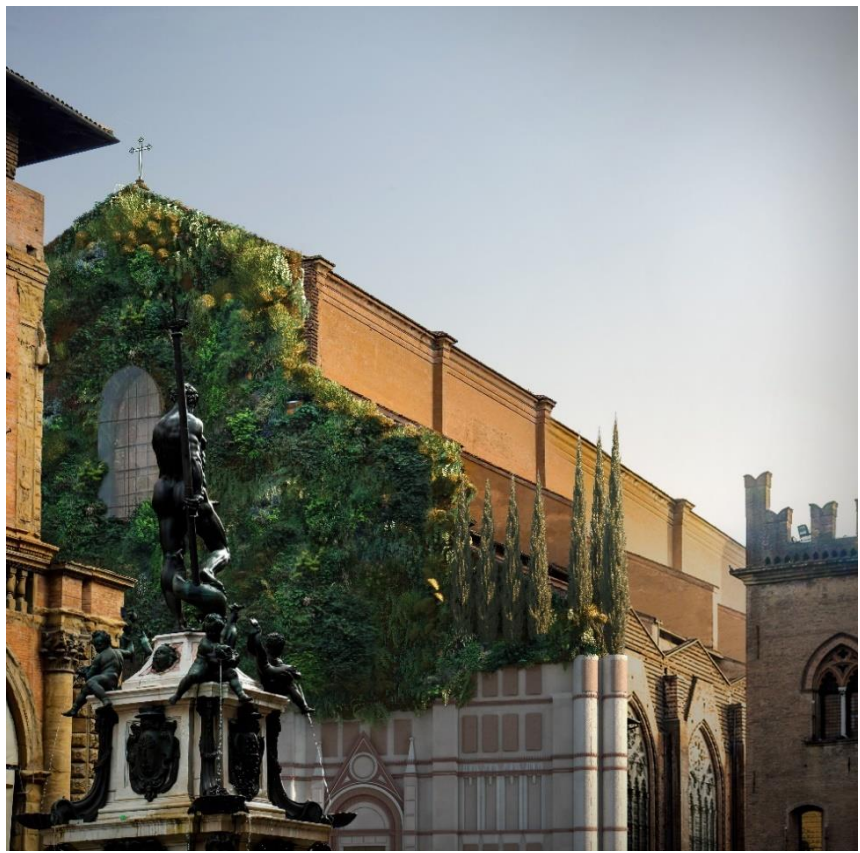

Figs. 3 and 4: *Bosco San Petronio*, Bologna, Italy. Mario Cucinella Architects. Images courtesy of Mario Cucinella Architects and MCA Visual

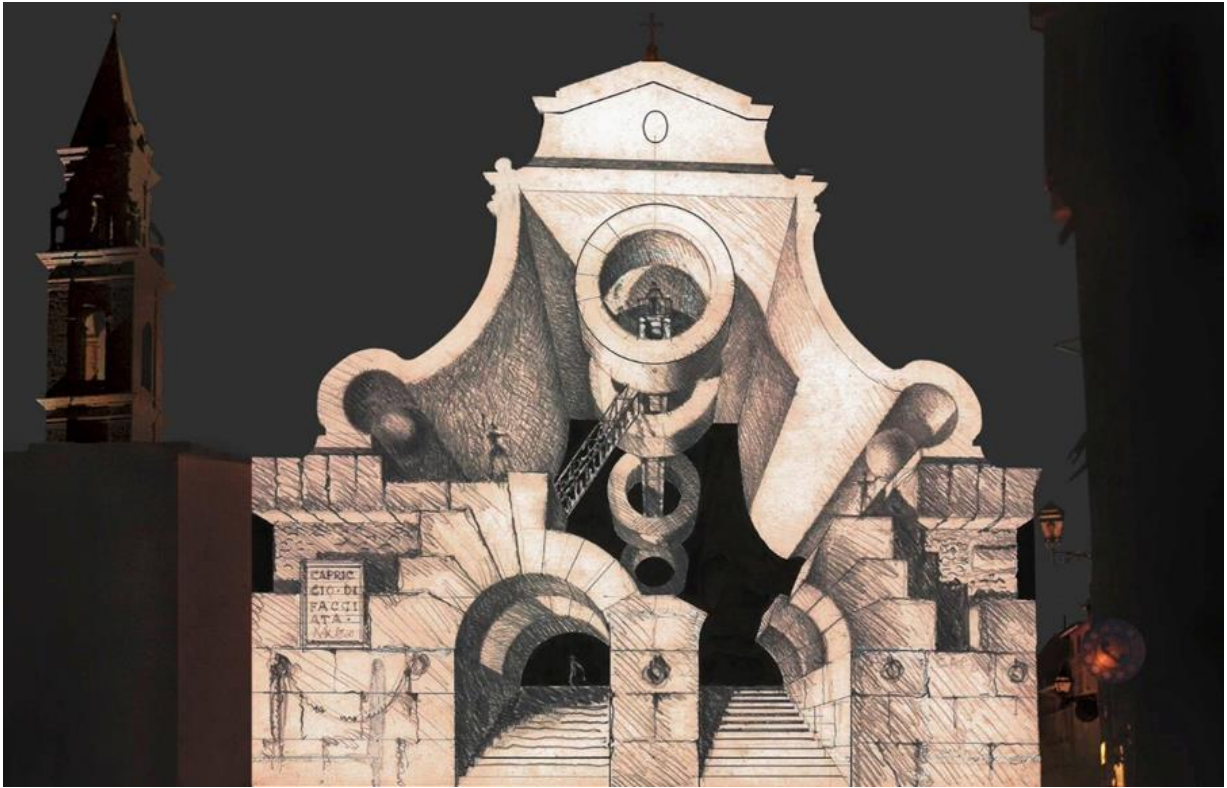

Fig. 5: Mario Mariotti (1980). Projection proposal for "il completamento della facciata del Santo Spirito". Project *Piazza della Palla*. Florence.

For the present investigation we will adopt some of the already classic examples of unfinished *quattrocento* façades in central Italy in order to examine how Artificial Intelligence operates for their completion (table 1). For observation, three situations have been distinguished: partially built façades (*imperfecta*), bare façades (*nuda*) and deferred façades (*perfecta dilata*), that is, those that were completed on a much later date than the original project.

For the study we will use DALL-E Outpainting tool [9] capable of extending images based on the pre-existing graphic elements, from whence it is possible to predict that it will use coherent visual resources to complete unfinished images. The purpose of this work is to explore the response of AI for different typologies of incompleteness of historic façades. The input is a 1024 x 1024-pixel image of the unfinished façade. It is necessary to accompany the image with a text, which will condition the result expected by the user. Once the texts and images have been processed, Outpainting outputs a solution, unique for each attempt, so that a different image is obtained each time, even when entering the same text sequence.

| Façade | Date of the façade | Typology |
|---|---|---|
| Badia of San Bartolomeo (Fiesole) | 1456. Façade cladding of the central doorway and two adjacent elements | IMPERFECTA |
| San Petronio (Bolonia) | 1425. Porta Magna<br><br>1524-1530. Façade cladding around adjacent doors to Porta Magna | IMPERFECTA |
| San Lorenzo (Florencia) | 1418. (Medicean phase) | NUDA |
| Santo Spirito (Florencia) | 1487 | NUDA |
| Santa Maria Novella (Florencia) | Uncomplete until 1470. Completed by L. B. Alberti. | PERFECTA DILATA |
| Santa Croce (Florencia) | Uncomplete until 1863. Completed by Niccolò Matas | PERFECTA DILATA |

Table 1. Analysed examples

Only the term "façade" has been introduced in the text sequence so as to provide the minimum information essential for observing the performance of the AI with the least possible interference. Two results are presented for each example, in order to examine the different variations and explore the degree of differentiation in consecutive trials, leaving the reader the possibility to replicate and amplify the investigation with additional trials.

### *IMPERFECTA PERFICERE: FINISHING THE UNFINISHED*

The examples chosen for the analysis of the completion of unfinished façades are the Badia Fiesolana (fig. 6) and San Petronio in Bologna (fig.7). In both cases, these are works that offer a visual guideline for the AI to have chromatic and geometric elements on which to base their reconstruction.

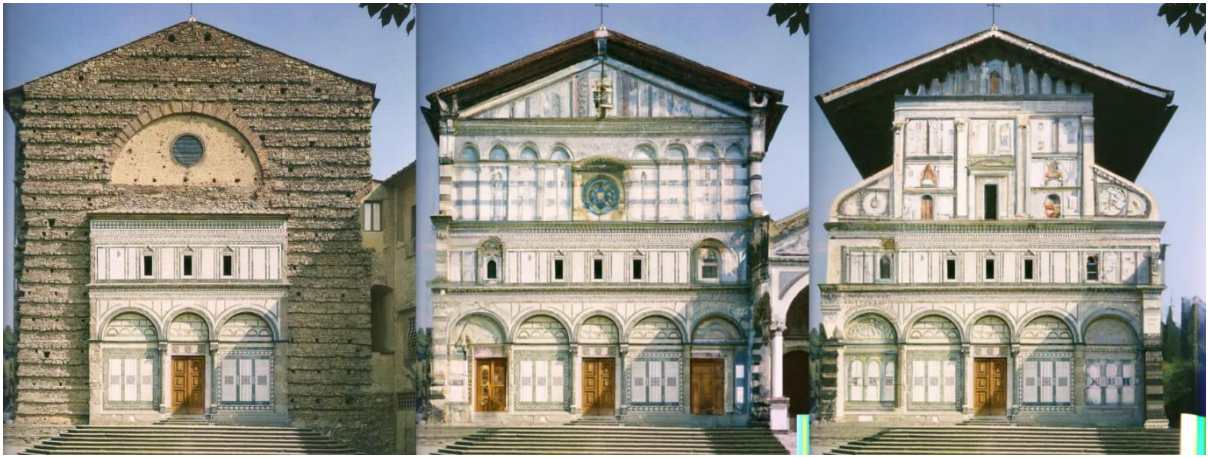

Fig. 6: Badia Fiesolana (Fiesole). From left to right: current state and two consecutive restoration proposals generated using DALL-E 2 Outpainting tool

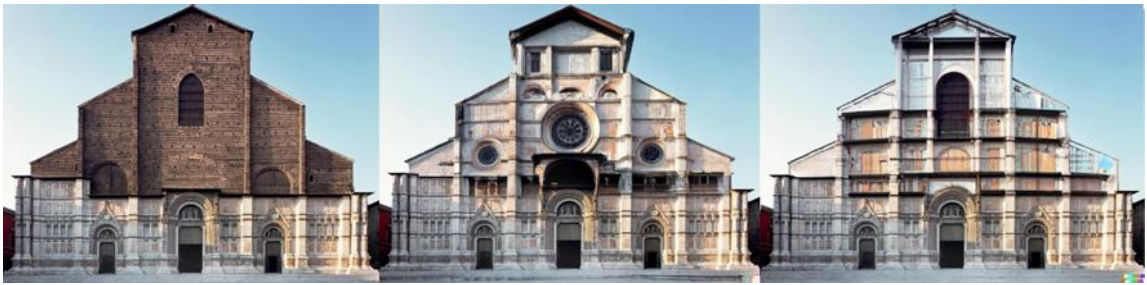

Fig. 7: Basilica of San Petronio (Bologna). From left to right: current state and two consecutive restoration proposals generated using DALL-E 2 Outpainting tool

It is observable that, in this and in all the cases examined, the AI respects the principle of symmetry according to the vertical central axis, typical of historical architectures up to the Modern Movement. The solutions appear, in very broad strokes, formally related to the architecture of the temples of central and northern Italy of the *quattrocento*. The decoration of the Badia Fiesolana, with the area's traditional cladding in white and serpentine marble from Prato, probably suggests to the IA the model of Santa Maria Novella, one of the most recognizable examples of the use of green and white bichrome on the façade. That is why in the second solution of fig. 6 appears an imitation of lateral volutes that undoubtedly recall the conformation of the Albertian façade of the Florentine basilica.

In the case of San Petronio, the AI operates with the freedom allowed by the chromatic limitations and produces solutions that are coherent with Romanesque or even Renaissance architecture, without echoing, at least in the solutions on display, the influence of Gothicist features, -non-Gothic-, from the side portals.

### *NUDA PERFICERE: FINISHING THE NAKED*

To examine how AI operates when there are no reference points other than the flat perimeter of the façade, two representative classic examples of such a situation have been chosen: the Basilica of San Lorenzo (fig. 8) and the Basilica of Santo Spirito (fig. 9), both in Florence.

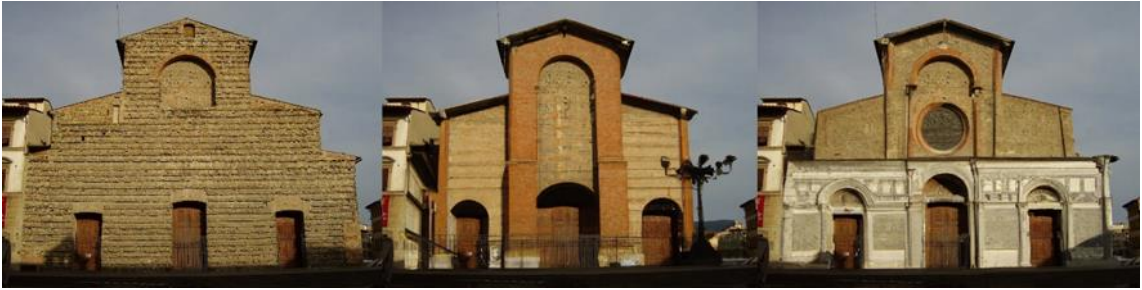

Fig. 8: Basilica of San Lorenzo (Florence). From left to right: current state and two consecutive restoration proposals generated using DALL-E 2 Outpainting tool

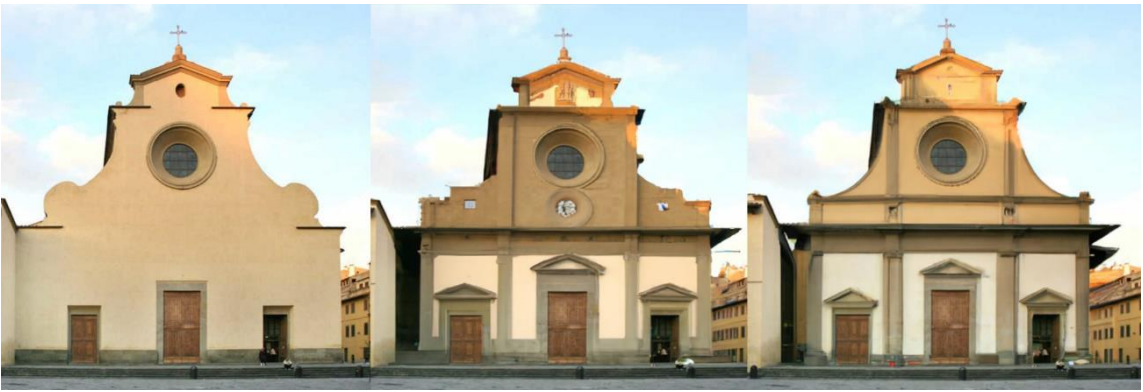

Fig. 9: Basilica of Santo Spirito (Florence). From left to right: current state and two consecutive restoration proposals generated using DALL-E 2 Outpainting tool

DALL-E 2 uses a generation process called "diffusion" that starts with a pattern of random dots and gradually alters that pattern into an image when it recognizes specific aspects of the provided image [10]. This operation is clearly shown in the façades lacking any kind of cladding: in the case of the examples analysed in this section, the AI determines that the inconclusiveness is characteristic of the starting image and returns solutions with clear features of that inconclusiveness. The issue becomes clear in San Lorenzo, in which the AI leaves part of the façade uncovered. In the case of Santo Spirito, the cloning of the façade render generates an apparently more finished example, since the tool naturally fails to perceive the originally provisional nature of the coating on the unfinished façade.

The absence of reference visual elements, other than the perimeter and the definition of openings, prevents the AI from identifying coherent architectural elements to choose a pattern and determines the results it offers for both basilicas. Despite the lack of definition of the images corresponding to the current state of the façades and the poor results obtained, in this case only the word "façade" has also been supplied, since the objective was to obtain an inventory of comparable solutions in the three case-study sections. In order to prevent the AI from identifying similar images on the web, the introduction of more defined text sequences, such as "Italian Renaissance church" or similar, from which the tool could have been induced to operate, has been deliberately omitted to achieve a complete reconstruction. This research has tried at all times to force the generation of solutions from graphic contextual elements.

In order to verify the operating limitations in these examples, a large number of consecutive attempts were made in addition to those presented here and, with different variations, the inconclusiveness in the case of San Lorenzo and the confusion caused by the current finishing mortar on the façade, in the case of Santo Spirito, were a constant in all the solutions offered by DALL-E-2.

## *PERFECTA REFICERE: REDOING THE FINISHED*

Finally, the analysis of façades whose completion took place at a time considerably after the completion of the building is proposed. Typical are some Florentine examples, among which Santa Maria Novella (fig. 10) and Santa Croce in Florence (figs. 11 and 12) stand out. More than a century elapsed between the construction of the temple of Santa Maria Novella and the Albertian design of its façade and more than four centuries mediated from the completion of the Santa Croce and the completion of its façade in the second half of the 19th century.

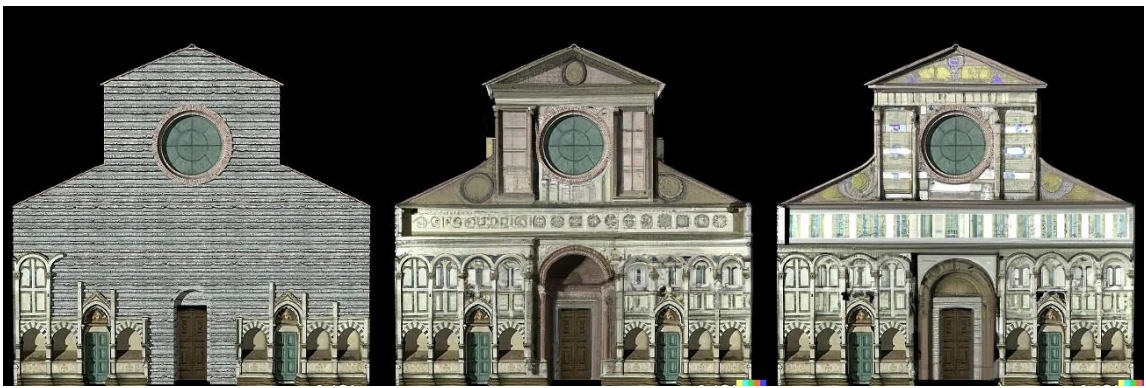

Fig. 10: Santa Maria Novella (Florence). From left to right: hypothesis of the state of the façade around 1350 [11] and two consecutive restitution proposals generated using the Outpainting tool of DALL-E 2

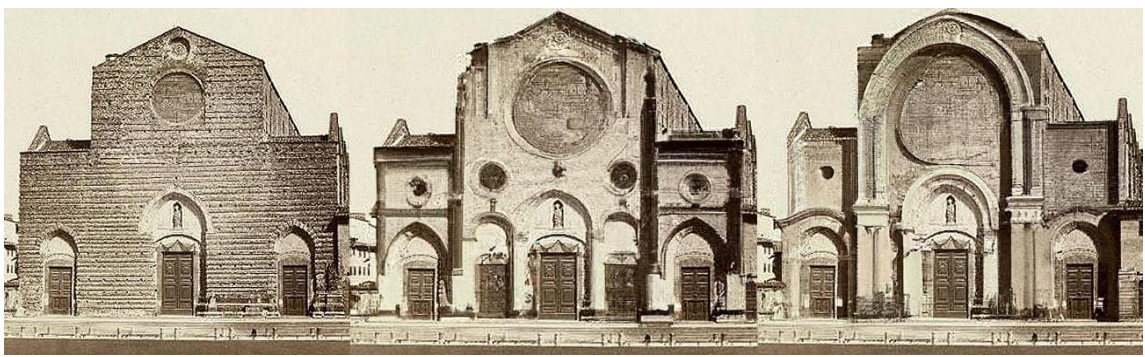

Fig. 11: Santa Croce (Florence). From left to right: state of the façade before 1853 and two consecutive restitution proposals generated using the Outpainting tool of DALL-E 2

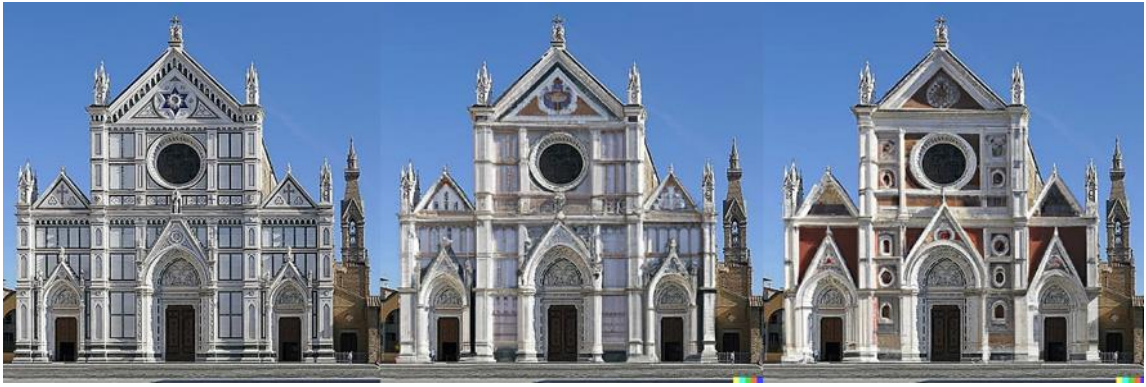

Fig. 12: Santa Croce (Florence). From left to right: present state of the façade (after 1853) and two consecutive restitution proposals generated using the Outpainting tool of DALL-E 2

The objective in this case is to observe the distance between the finished façades, albeit deferred in time, and the proposals generated by the AI, in order to compare them with the proposals of their natural authors.

For Santa Maria Novella, we start from the hypothesis of the state of the façade before the intervention of León Battista Alberti according to a reconstruction by 3dSign Studio [11]. Outpainting returns correctly based proposals that are somewhat reminiscent of the real façade designed by Alberti. The pattern defined by the pre-existing allows the AI to formulate an acceptable level of consistency in the creation of its proposals.

In the case of Santa Croce, a double observation has been made. Since photographic documentation of the uncoated façade is available, it has been input in this way in the tool, generating a result similar to that already observed for San Lorenzo. Outpainting detects the inconclusiveness and offers proposals that, in turn, also appear unfinished.

For the second attempt, a contemporary photograph is inserted into the tool in which the covering equivalent to the state that the façade had before 1853 is erased, defining only the perimeter and the openings of the Santa Croce. The results of the AI are consistent with the natural result of the current façade. The hypothesis is that the AI has identified the Baccani campanile, in the background on the right, and has based the reconstruction on sets of current images of Santa Croce. This is demonstrated by the fact that when the campanile is removed to launch new attempts, the AI begins to distance itself greatly from the real façade (fig. 13).

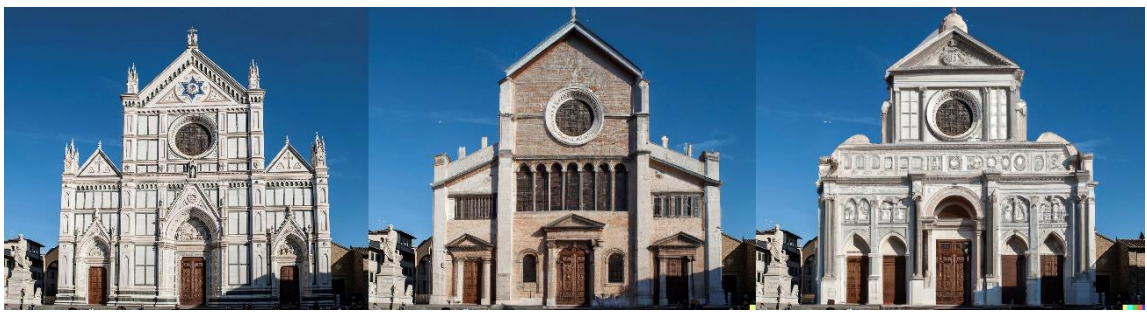

Fig. 13: Santa Croce (Florence). From left to right: state of the façade in the present, cutting out the characteristic campanile, and two consecutive restoration proposals generated by the Outpainting tool of DALL-E 2

*CONCLUSIONS*

The results generated by the DALL-E-2 Outpainting tool for the completion of unfinished facades are generally consistent with what is expected of a photorealistic reconstruction based on the graphic resources that the AI is able to identify on the web.

The estrangement in the solutions proposed by the AI is directly proportional to the limitation of the visual contextualization data that it is capable of identifying. The natural proposals that throughout history have been raised by many architects for some of the facades analyzed show greater independence in this regard. Thus, facades with defined starting patterns, such as the Basilica of San Petronio in Bologna, have given rise to contained and coherent projects in different competitions and also to proposals of creative frenzies of arguable formal foundations.

In this sense, it can be said that the architect is creative, for better or for worse, when he wants and the AI when it has no other resources. From this observation it could be derived that AI is more objective and, therefore, more scientific to the extent that it has a more regular way of operating, however, it is evident that some of the proposals presented here can be discarded based on stylistic, historical, formal or aesthetic criteria (e.g. figs. 8, 11 and 13).

Despite the embryonic state of the tool, and considering that its purpose is to create new images based on verbal and visual contexts, it is possible to predict that the results will improve along with future improvements in the identification of architectural elements and compositional rules within historical visual contexts, which, perhaps could lead to more consistent propositional examples. At the moment, it is an instrument that can be classified more in the field of mere artistic creativity than in the discipline of Architectural Theory.

*NOTES*

1/ "[…] propter interpellationem reipublicae incepta reliquerunt". (Vitruvio, 1997, p. 169)

2/ "Athenians […] deliberately left the ambitious Temple of Olympian Zeus unfinished as a memorial to tyrannical hubris". (Stuttard, 2021).

3/ "[…] In these times, due to lack of devotion or to the greed of men, a large church is no longer started, nor are to be finished those already started". Translated by the authors.

4/ On the referendum in Florence for the completion of Santa Maria del Fiore, in the 13th century (Saalman, 1964, p. 490); on the 1872 Santa Maria del Fiore contest (Cresti, Cozzi, Carapelli, 1986, p. 169); on the 1886 competition for San Petronio: […] il Comitato […] ebbe la cattiva idea di indire un referendum popolare per decidere quale dei tre progetti giudicati a pari merito fosse realmente il migliore (Piacentini, 1935, p. 401). Despite this, in 1933 a new contest will be called; on the proposal to finish the Basilica of San Lorenzo in Florence in the 21st century (Repubblica, 2011).

5/ "Completare la basilica di San Lorenzo di Firenze, costruendo la facciata in marmo, mai realizzata e progettata da Michelangelo nel 1515: una proposta da sottoporre a un referendum tra i fiorentini. Per prendere una decisione entro il 2015: a 500 anni dal progetto di Michelangelo e a 150 da Firenze Capitale".  "Completing the Basilica of San Lorenzo in Florence by building the marble façade designed by Michelangelo in 1515 and never carried out: a proposal to be submitted to referendum among the Florentines to adopt a decision before 2015: when the 500th anniversary of Michelangelo's project and 500 since Florence became the capital" (Repubblica, 2011). (Translated by the Authors).

The proposal, predictably, was discarded before even submitting it to a referendum among the Florentines.

6/ "Restaurer un édifice, ce n'est pas l'entretenir, le réparer ou le refaire, c'est le rétablir dans un état complet qui peut n'avoir jamais existé à un moment donné". (Viollet-le-Duc, 1866, p. 14)

7/ "aucune civilisation, aucun peuple, dans les temps écoulés, n'a entendu faire des restaurations comme nous les comprenons aujourd'hui". (Viollet-le-Duc, 1866, p. 14)

8/ "Do not let us deceive ourselves in this important matter; it is impossible, as impossible as to raise the dead, to restore anything that has ever been great or beautiful in architecture." (Ruskin, 1889, p. 194).

9/ https://openai.com/blog/dall-e-introducing-outpainting/

10/ < https://openai.com/dall-e-2/>

11/ Digital reconstruction made for the exhibition *L'uomo del Rinascimento. Leon Battista Alberti e le arti a Firenze tra ragione e bellezza* (Firenze, 11 marzo-23 luglio 2006), in: https://www.youtube.com/watch?v=i3FYkjcY4uA&t=172s

---

We want to thank Prof. David Stuttard for his kind contribution to this research in the section on the inconclusiveness of classical buildings.